\documentclass[letterpaper]{article} 
\usepackage{aaai24}  
\usepackage{times}  
\usepackage{helvet}  
\usepackage{courier}  
\usepackage[hyphens]{url}  
\usepackage{graphicx} 
\usepackage{natbib}  
\usepackage{caption} 
\usepackage{algorithm}
\usepackage{algorithmic}
\usepackage{amsfonts} 
\usepackage{amsmath}
\usepackage{color}
\usepackage{xcolor}
\usepackage{newfloat}
\usepackage{listings}
\DeclareCaptionStyle{ruled}{labelfont=normalfont,labelsep=colon,strut=off} 
\floatstyle{ruled}
\newfloat{listing}{tb}{lst}{}
\floatname{listing}{Listing}
\nocopyright 

\newcommand{\papername}{StyleBrush}

\title{\papername: Style Extraction and Transfer from a Single Image}
\author{
    \small
    Wancheng Feng\textsuperscript{\rm 1}  
    , Wanquan Feng
    , Dawei Huang\textsuperscript{\rm 1}
    , Jiaming Pei\textsuperscript{\rm 1 2}
    , Guangliang Cheng\textsuperscript{\rm 1 3}
    , Lukun Wang\textsuperscript{\rm 1}\thanks{Corresponding author.}\\
}
\affiliations{
    \textsuperscript{\rm 1}Shandong University of Science and Technology, Taian Shandong China\\
    \textsuperscript{\rm 2}University of Sydney, Sydney Australia\\
    \textsuperscript{\rm 3}Liverpool University, Liverpool United Kingdom\\
    fengwancheng@sdust.edu.cn, wanquan0322@gmail.com, \\huangdawei@sdust.edu.cn, jpei0906@uni.sydney.edu.au, \\
    Guangliang.Cheng@liverpool.ac.uk, 
    wanglukun@sdust.edu.cn 

}

\usepackage{bibentry}

\begin{document}

\maketitle

\renewcommand{\thefootnote}{}
\footnotetext{Under Review.}


\begin{center}
    \includegraphics[width=8.5cm]{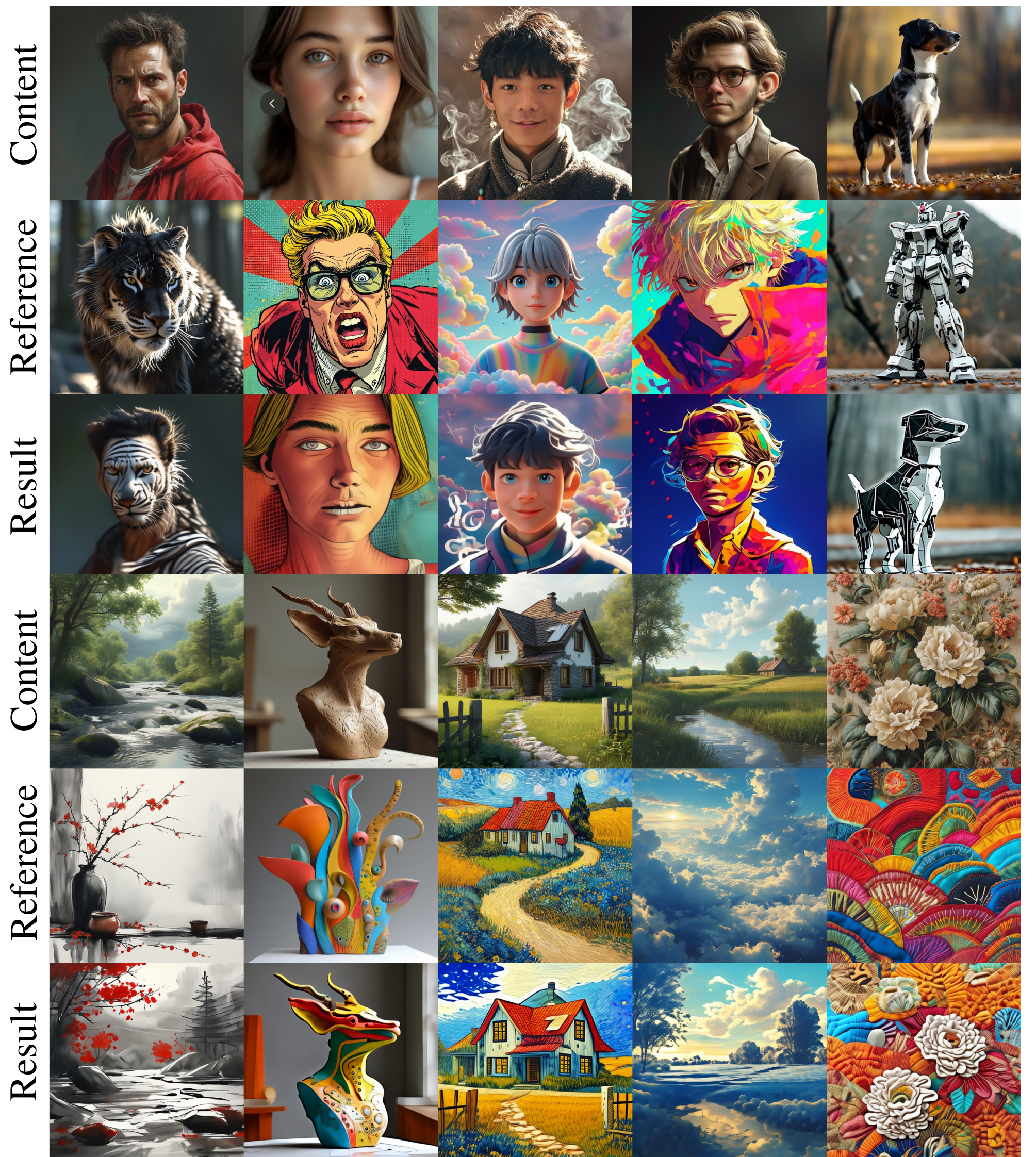}
    \captionof{figure}{We propose \textbf{\papername}, a framework that transfers style from only a single reference style image. The gallery above shows our results on various styles and diverse content images. The $1,4-th$ rows are the contents, the $2,5-th$ rows show the styles, and the $3,6-th$ rows are the results.}
    \label{fig:teaser}
\end{center}


\begin{abstract}
Stylization for visual content aims to add specific style patterns at the pixel level while preserving the original structural features.
Compared with using predefined styles, stylization guided by reference style images is more challenging, where the main difficulty is to effectively separate style from structural elements.
In this paper, we propose \textbf{\papername}, a method that accurately captures styles from a reference image and ``brushes'' the extracted style onto other input visual content.
Specifically, our architecture consists of two branches: ReferenceNet, which extracts style from the reference image, and Structure Guider, which extracts structural features from the input image, thus enabling image-guided stylization. We utilize LLM and T2I models to create a dataset comprising $100K$ high-quality style images, encompassing a diverse range of styles and contents with high aesthetic score. To construct training pairs, we crop different regions of the same training image. 
Experiments show that our approach achieves state-of-the-art results through both qualitative and quantitative analyses. We will release our code and dataset upon acceptance of the paper.
\end{abstract}

\section{Introduction}

Visual content encompasses a vast array of styles, drawing from various artistic disciplines and the unique perspectives of different artists. Stylization~\cite{bruckner2007style, deng2022stytr2, chen2021artistic, wang2023styleadapter} is the process of applying a specific artistic style to content. Stylization play an important role in generative AI, and serves as a valuable tool across various domains, including artistic creation, film production, entertainment, and even digital marketing.

There are various style representation modalities. One strategy is through descriptive text, such as style labels like ``oil painting'' or ``anime style''. However, text prompt tends to be brief and coarse. Alternatively, styles can be pre-defined or preset, such as with methods like DreamBooth~\cite{ruiz2023dreambooth} or LoRA~\cite{hu2021lora}; however, these approaches are resource-intensive as they require individual training for each style. Considering these limitations, in this work, we explore the utilization of reference images as style representations, which achieves a harmonious balance between user convenience and the level of style representation details.
There are two critical components when transferring style from a reference image to a content image. The first key component is the accurate extraction of the style elements from the style image. The second, equally crucial, component is preserving the structural integrity of the content image throughout the stylization process. The goal is to find a delicate balance so that the style can be transferred while retaining the original structure.

To achieve a satisfactory stylization effect, we need to overcome several critical challenges. The first major challenge is to design an appropriate architecture that can effectively separate style and structure from the input images. The second challenge is to design a method that utilizes priors from pre-trained visual generation models~\cite{rombach2022high}, which avoids the need to train an expert model from scratch and helps ensure high-quality output. The third challenge is to reduce time and resource consumption by avoiding optimization during inference and using only the forward pass for stylization.

Although traditional texture synthesis-based style transfer methods~\cite{bruckner2007style} can achieve correct stylization effects, we do not refer to this strategy because of its huge time consumption. In contrast, some transformer-based stylization models~\cite{deng2022stytr2, chen2021artistic} are data-driven, but their generation effects are limited because they do not utilize the prior knowledge of large visual models. Some recent~\cite{chung2024style} training-free methods based on pre-trained large models effectively utilize the prior knowledge, but are limited by the hand-designed feature calculation methods. In addition, some style transfer methods~\cite{wang2023styleadapter, qi2024deadiff} are trained to embed styles but lack the capacity to analyze content images. To apply them to stylization tasks, they require supplementary structural preservation schemes, like ControlNet~\cite{zhang2023adding}, ultimately diminishing their control over maintaining structures effectively.

In this paper, we propose \textbf{\papername}, a framework that can elegantly address these challenges. We employ a two-branched structure based on pre-trained large model as our model structure. Specifically, we are inspired by the previous method Animate Anyone~\cite{hu2023animate}, which can drive a character to dance. The input includes a reference image to provide human texture information and a pose image to provide structure information, and finally generates a result that obeys the input texture and structure. In our task, the style image is similar to the reference character appearance image in Animate Anyone, and the content image is similar to the human pose image, thus establishing a delicate logical correspondence. Naturally, we imitated the network structure of Animate Anyone, design a Structure Guider to encode the structure of the content image, and use a Referencenet to encode the style. In order to ensure that what the structure guider learns is structure and does not completely contain style, we process the content image into a grayscale image to remove the color style and blur it to remove the line style.

During training, we utilize the public Wikiart dataset~\cite{artgan2018} along with a specifically constructed dataset to train our model. For the construction of our dataset, we generate text prompts using GPT-4o~\cite{artgan2018}, produce images with Kolors~\cite{kolors}, and assess aesthetic scores using Q-align~\cite{wu2023qalign}. The dataset comprises a total of $100K$ images. From any given single training image, we can crop two different areas to create a pair for training. Our trained model effectively extracts styles from reference images, maintains structural integrity in content images, and generates sophisticated results. Thanks to Animatediff~\cite{guo2023animatediff}, our method can be extended to video stylization at no additional cost. Our experimental results demonstrate leading performance in both qualitative and quantitative metrics.

In summary, the \textbf{contributions} of our study can be enumerated as follows:

\begin{itemize}
    \item We propose a novel framework for reference image-guided stylization that effectively decouples style and structure from the input images.
    \item We develop a stylization training process using a data pipeline that constructs training pairs from single images.
    \item We construct a dataset containing $100K$ high-quality style images, encompassing a diverse range of styles and contents with high aesthetic score.
    \item We achieve satisfactory stylization effects, demonstrated through both quantitative and qualitative analyses.
\end{itemize}

\begin{figure*}[t]
    \centering
    \includegraphics[width=17.5cm]{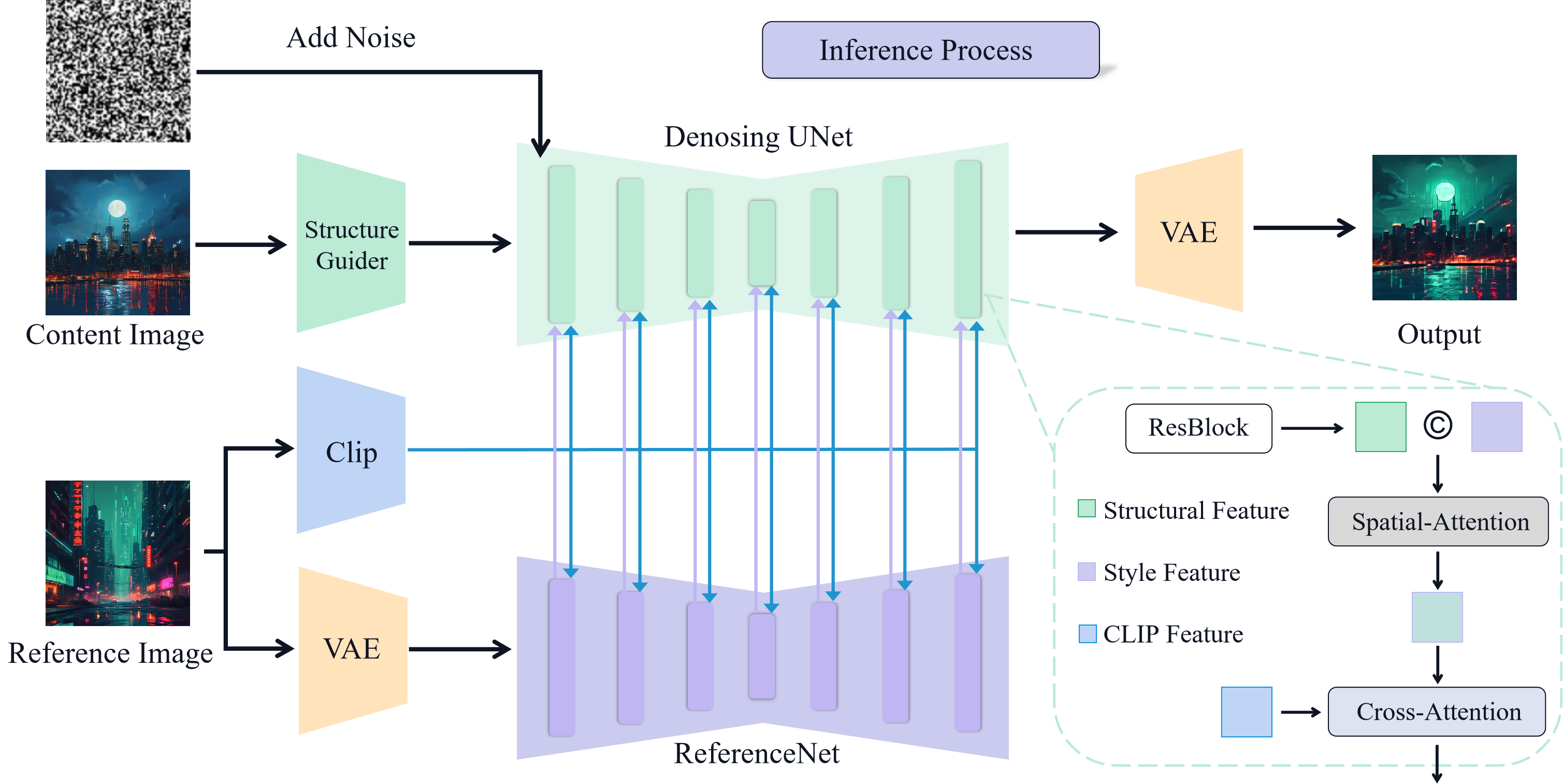}
    \caption{Overall of \textbf{\papername} pipline. The Structure Guider processes the content image, while the reference image is handled by the CLIP encoder and VAE encoder modules. Noise is added to the content image, which is then passed through the Denoising UNet along with features from the ReferenceNet. ReferenceNet integrates style and CLIP features. Finally, the combined features in the Denoising UNet are passed to the VAE decoder to generate the output image, effectively blending the structure and style of the input images.
    }
    \label{fig:main}
\end{figure*}

\section{Related Work}

{\noindent \bf Diffusion-based Image Generation} 

In recent years, diffusion models have made remarkable strides in the field of image generation. In 2020, the introduction of DDPM \cite{ho2020denoising} marked a significant advancement by employing Markov processes to address denoising tasks. That same year, DDIM \cite{song2020DDIM} further evolved the approach by removing the Markov chain constraint, thereby reducing the sampling steps and simplifying the inference process. In 2021, Dhariwal \cite{dhariwal2021diffusion} showcased that diffusion models surpassed GANs in generative performance. The results for LDM \cite{rombach2022high} and stable diffusion (SD) are impressive. Despite these advances, achieving controllability over the source image during generation remains a challenge. In the last two years, the strong prior knowledge embedded in open-source pre-trained stable diffusion models has led to the increasing adoption of this architecture for image generation tasks \cite{ruiz2023dreambooth, gal2022image, hu2021lora}. With the introduction of ControlNet \cite{zhang2023adding}, controllable generation in image tasks has become a reality. Today, diffusion model-based generation stands as the most mainstream approach in the field.

{\noindent \bf Image Guided Stylization} 

Image Guided Stylization has been a well-established topic in computer vision, with research tracing back to the mid-1990s \cite{jing2019neural}. Early works in stylization \cite{bruckner2007style} often utilized brushstroke techniques to achieve stylization effects. Between 2016 and 2017, seminal approaches \cite{gatys2016image, ulyanov2016instance} laid the foundation for modern stylization techniques, though these models were initially limited to transferring a single style image at a time. To overcome this limitation, Golnaz et al. \cite{ghiasi2017exploring} introduced a style prediction network, enabling a single model to perform stylization across multiple styles. In 2022, Deng et al. \cite{deng2022stytr2} demonstrated the superior performance of Transformers in stylization tasks, marking a significant advancement in the field. Since 2023, diffusion model-based stylization has gained prominence. Notable examples include adapter-based methods like StyleAdapter \cite{wang2023styleadapter} and InstantStyle \cite{wang2024instantstyle}; inversion and shared attention-based methods like StyleAlign \cite{wu2021stylealign} and Visual Style Prompting \cite{jeong2024visual}; as well as test-time fine-tuning approaches such as StyleDrop \cite{sohn2023styledrop}. These methods typically leverage the extensive prior knowledge of diffusion models to understand and manipulate image structure and artistic elements.

\section{Method}

\begin{figure}[t]
    \centering
    \includegraphics[width=8.5cm]{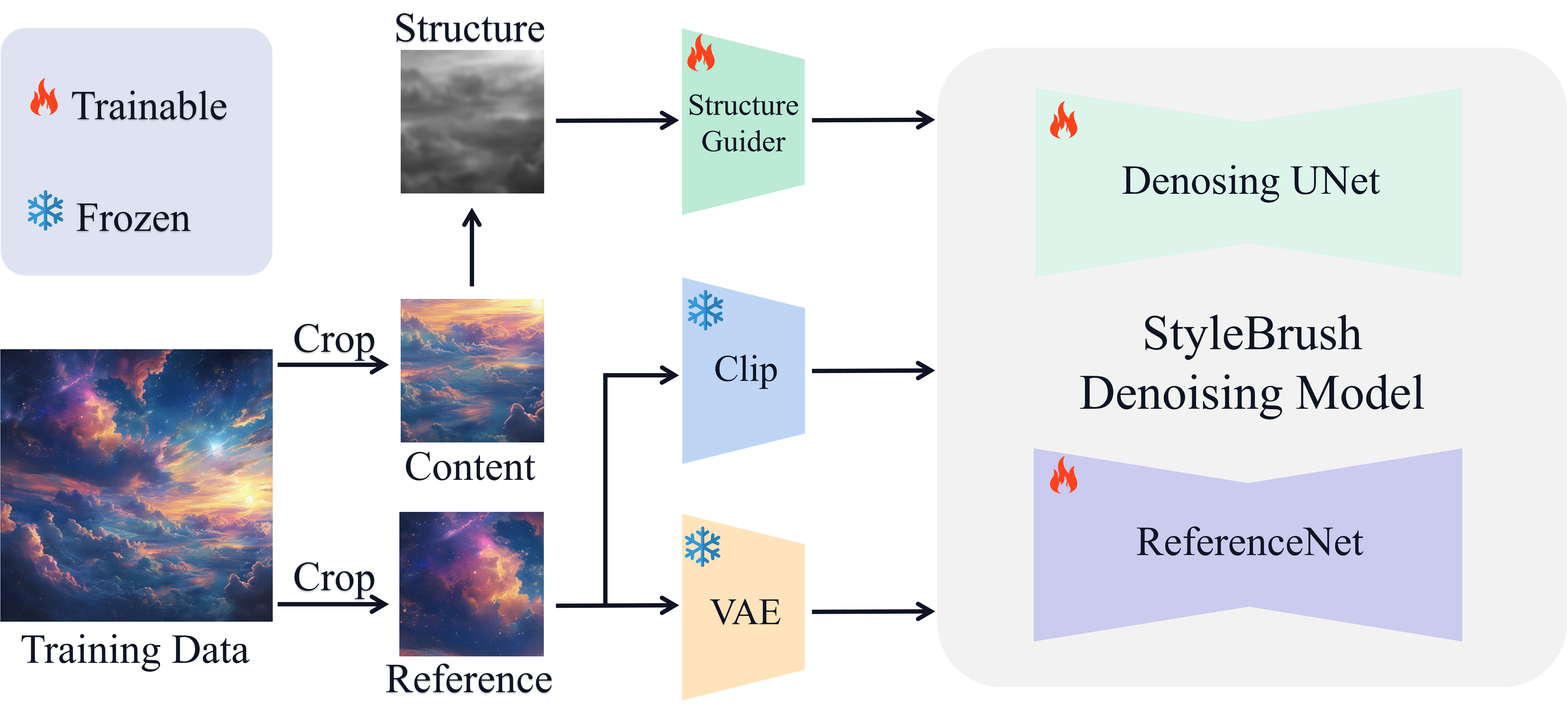}
    \caption{\textbf{\papername} training process. The images are randomly cropped to generate both content and reference images. The content image is then transformed into a structural image and input into the trainable Structure Guider module. The reference image is processed separately using frozen CLIP and VAE models. The StyleBrush denoising model combines these inputs to produce the final denoised image. Modules marked with a flame icon are trainable, while those marked with a snowflake icon remain frozen during training.}
    \label{fig:train}
\end{figure}

In this paper, we introduce \textbf{\papername}, a reference image guided stylization method based on diffusion model. We firstly introduce some background knowledge about our base model, stable diffusion, in the section of preliminaries. Then we describe our stylization pipeline and the training strategy in the following sections. The pipeline and training process are illustrated in Figure~\ref{fig:main} and Figure~\ref{fig:train}, respectively.



\subsection{Preliminaries: Stable Diffusion}
\label{subsec:3.1}
In this section, we provide a introduction to Stable Diffusion~\cite{rombach2022high}. 
Diffusion models exhibit strong performance in image and video generation and editing tasks. In this work, we adopt Stable Diffusion as our base model. Initially, the content image $x_0$ is fed into the latent encoder, resulting in $z_0=\mathcal{E}(x_0)$. Gaussian noise is then added to the latent. The forward process of diffusion can be described as follows:

\begin{equation}
q(z_t|z_{t-1})=\mathcal{N}(z_t;\sqrt{1-\beta_t}z_{t-1},\beta_tI),
\label{con:addnoise}
\end{equation}
where ${\beta}_t(t=0,1,... ,T)$ represents the scale of noise. $T$ denotes the number of steps in the forward diffusion process. After obtaining the latent noise, it is fed into the UNet. The backward noise reduction step of UNet, from $z_t$ to $z_{t-1}$, can be described as follows:  

\begin{equation}
p_\theta(z_{t-1}|z_t)=\mathcal{N}(z_{t-1};\mu_\theta(z_t,t),\Sigma_\theta(z_t,t)),
\end{equation}
where $\mu_\theta$ and $\Sigma_\theta$ are implemented using a denoising model $\epsilon_\theta$ that has been trained on example data. During training, the denoising matching score is optimized according to the following:

\begin{equation}
L=\mathbb{E}_{\mathcal{E}(z),\epsilon\sim \mathcal{N}(0,1),t}\left[{||\epsilon-\epsilon_\theta(z_t,t)||}^2_2\right].
\end{equation}
Once the model is trained, the inference process proceeds as follows: Noise is initially added to the latent, and the image is subsequently denoised using the DDIM procedure, which predicts the value of $z_{t-1}$. Finally, the denoising result is fed into the latent decoder $\mathcal{D}(\cdot)$, which produces the output image $\hat{x_0} = \mathcal{D}(\hat{z}_0)$.

\subsection{Stylization Pipeline\label{subsec:3.2}}

We show the whole pipeline in Figure~\ref{fig:main}. Inspired by recent work on human dance~\cite{hu2023animate}, virtual try-on and human image generation~\cite{choi2024improving,chen2024wear,huang2024parts}, we refer to their design and utilize a two-branched structure to deal with the texture feature and structure feature, respectively.
As Figure~\ref{fig:main} shows, our framework consists of two main branches: the ReferenceNet branch and the Denoising UNet branch. Both branches use the same UNet architecture and leverage the pre-trained model of Stable Diffusion. The ReferenceNet branch extracts pixel level features from the input image, while the Denoising UNet incorporates structural information from the Structure Guider.


To extract a clean structure from the content image, we first convert the image to grayscale and then apply a blur. This approach removes texture details while preserving the structural elements, as illustrated in Figure~\ref{fig:train}. By eliminating the color features of the content image and effectively removing its structural and textural information, we create a ``canvas'' for stylization. The processed image is then fed into the Structure Guider as input for Denoising UNet. We use the same $4\times4$ kernel and $2\times2$ strides as in \cite{zhang2023adding}. The input image is processed through four convolutional layers with channels of $(16, 32, 64, 256)$ before being fed into the Denoising UNet, thereby preserving the structure level features of the image.

\begin{figure*}[t]
    \centering
    \includegraphics[width=17.5cm]{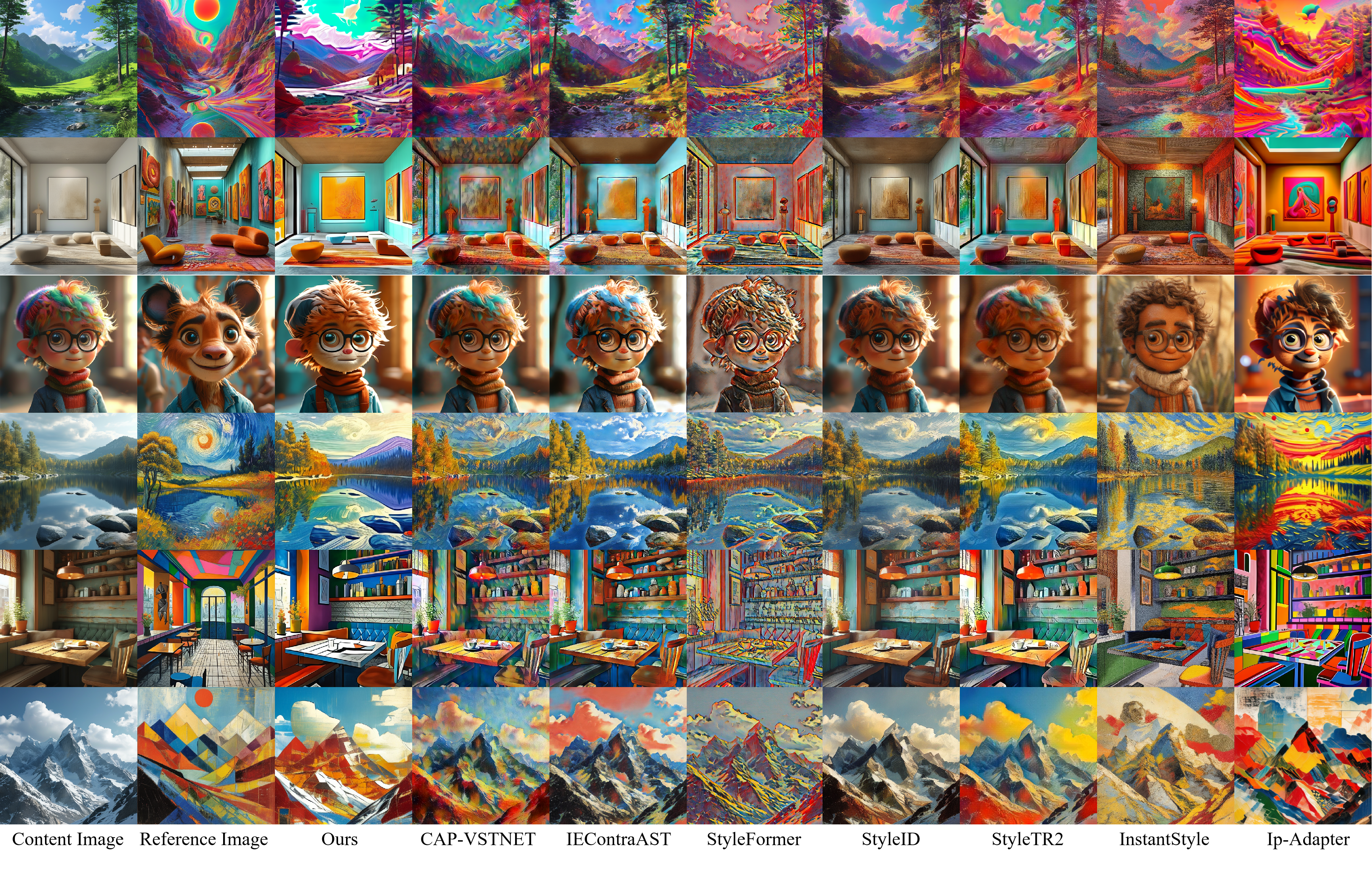}
    \caption{The qualitative comparison between our method and previous approaches, including CAP-VSTNET, IEContraAST, StyleFormer, StyleID, StyleTR2, InstantStyle, and IP-Adapter. As shown in the figure, our method achieves satisfactory stylization effect while maintaining the structure well.}
    \label{fig:4}
\end{figure*}

To optimize the extraction of stylistic features from the reference image, we observe that ReferenceNet maintains impressive pixel level coherence. ReferenceNet utilizes the same framework as the Denoising UNet but replaces its self-attention layer with a spatial attention layer. This adjustment enables ReferenceNet to inherit and update the original weights from Stable Diffusion during training. In our approach, we first encode the reference images using a VAE and then process them through ReferenceNet, ensuring precise extraction of stylistic elements. The feature map from the Denoising UNet is represented as $x_1 \in \mathbb{R}^{h \times w \times c}$, while the feature map from ReferenceNet is represented as $x_2 \in \mathbb{R}^{h \times w \times c}$. We concatenate $x_2$ along the width dimension with $x_1$, apply self-attention, and extract the first half of the feature map as the output.

Through these two structures, our method effectively captures both the structure level and style level features of the image. Similar to other stylization approaches \cite{zhang2023inversion}, we integrate a CLIP \cite{radford2021learning} encoder that shares its feature space with the VAE encoder. This shared space enhances semantic guidance throughout the generation process. We perform cross-attention between the structure level features of the content image and the style features of the reference image. These design choices enable our method to skillfully apply the style ``brush'' to the ``canvas''.

For video input, we integrate an additional temporal layer into the image stylization model, following the approach in \cite{guo2023animatediff}. This temporal layer is added after the spatial-attention and cross-attention mechanisms within the Res-Trans block, ensuring that it does not impact the image stylization effect. To be specific, we put a feature map $x{\in}\mathbb{R}^{b{\times}t{\times}h{\times}w{\times}c}$ reshape $x{\in}\mathbb{R}^{(b{\times}h{\times}w){\times}t{\times}c}$ before running temporal attention. In this part, we directly utilize the open-source temporal consistency method without requiring additional training \cite{MooreAnimateAnyone}.

Additionally, to ensure that our approach can preserve the image structure according to varying user preferences, we introduce a style strength parameter. 
To conduct the style strength adjustment strategy, we first replace the style image with the content image, perform a denoising process to obtain a latent (we denote as $latent_{c}$), and then linearly interpolate it with the original stylized latent (we denote as $latent_{s}$). The interploted latent can be computed as: $latent_{s} \cdot strength + latent_{c} \cdot (1-strength)$.

\subsection{Training Strategy\label{subsec:3.3}} 

In addition to designing the StyleBrush pipeline, we also establish a training strategy to train our stylization framework, as illustrated in Figure \ref{fig:train}. The training focuses on three main components: the style feature extraction capability of ReferenceNet, the structural feature extraction capability of Structure Guider, and the stylization capability of the Denoising UNet. For stylization training, we start with an input training image. Then we generate the content image and reference image by randomly cropping the content image. The content image is then processed to grayscale and blurred to serve as the structural representation, while the ground truth remains the content image.We initialize the Denoising UNet and ReferenceNet using pre-trained weights from Stable Diffusion. Structure Guider is initialized with Gaussian weights, and only the final projection layer uses zero convolutions. The weights for the encoder and decoder of the VAE are kept fixed. Our objective is to transfer the style of the reference image to the content image. To achieve this, we employ the following training strategy:

\subsubsection{Random Crop} 
To assess the training effectiveness, an intuitive approach is to use the structure image as input and employ its own style to reconstruct the content image. However, if the content image is directly used as the reference image to provide color information during training, the model will inevitably learn the structure information of the image itself. Given that the style of an image is generally consistent, we implement a random cropping strategy to prevent ReferenceNet from learning the structural information of the content image. In this approach, for a set of input images, we perform five random cropping attempts to identify two center points that yield the most distinct results. By comparing the distances between each pair of cropped images, we select the pair with the largest distance as the final content image and reference image.

\subsubsection{Structure Extraction} 
When handling the structural representation of images, both excessive preservation and destruction of structure can adversely affect the stylization task. Excessive preservation restricts the transfer of texture details from the style, while excessive destruction results in the loss of crucial structural information from the content image. In this paper, we adopt a simple yet effective method for structure representation. We firstly convert the content image into a grayscale version, and then apply a blurring operation to it, which removes texture details while retaining the basic structural elements. To avoid unwanted regularities associated with any specific blurring operration, we employ a random nested blurring strategy, which involves a three-layer nested approach combining MinFilter, GaussianBlur, and BoxBlur.

\section{Experiment}

\subsection{Settings}

In this section, we describe the implementation details, the datasets, and the evaluation metrics used in our experiments.

\subsubsection{Implementation Details}
We use Stable Diffusion v1-5~\cite{SD1.5} as the base model. The training processes is completed on $8$ A40 GPUs. We train with batch size $8$ for $30K$ iterations. A full inference run on a $512 \times 512$ image takes about $4$s on a A40 GPU.

\begin{figure*}
    \centering
    \includegraphics[width=17.5cm]{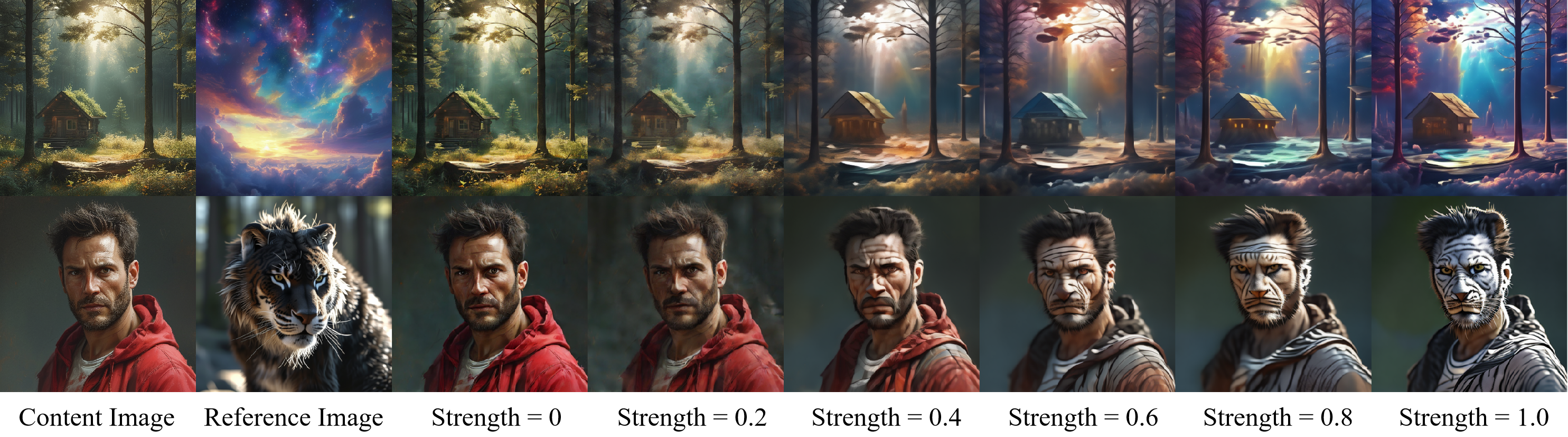}
    \caption{Results for various style strength values, ranging from $[0, 1]$. When the style strength is close to 0, the result is very similar to the content image; when the style strength is close to 1, the style of the result becomes consistent with the reference image.}
    \label{fig:5}
\end{figure*}


\subsubsection{Datasets}
We train our model on the \textbf{Wikiart} dataset~\cite{artgan2018} and a new dataset that we construct using GPT-4o~\cite{openai2024} and Kolors~\cite{kolors}, namely the StyleBrush dataset. For the Wikiart dataset, we divide it into the training set and the testing set. The number of images in the training set is $80944$, while the number of the testing images is $500$. For the StyleBrush dataset, we conduct a data generation pipeline. We first generate prompts for various artistic styles and contents (e.g. Chinese ink painting, modern arts, Japanese anime), using GPT-4o~\cite{openai2024}. These prompts are then used to create images with Kolors~\cite{kolors}, with each prompt generating four images from different random seeds. To ensure the high aesthetic quality of training images, we apply the aesthetic score of Q-Align~\cite{wu2023qalign} to filter out low-quality images. The final dataset consists of a total of $100K$ images.

\begin{table}[h!]
  \begin{center}
    \caption{For quantitative analysis with other stylization works, we use four metrics: ArtFID$\downarrow$, FID$\downarrow$, LPIPS$\downarrow$, and User$\uparrow$. ArtFID$\downarrow$ serves as the overall stylization metric, while FID$\downarrow$ and LPIPS$\downarrow$ assess style similarity and structural fidelity, respectively, with lower values indicating better performance. User$\uparrow$ represents user ratings, with higher values indicating better user preference. 
    }
    \begin{tabular}{l|c|c|c|c} 
      \hline
      \textbf{Methods} & \textbf{ArtFID$\downarrow$} & \textbf{FID$\downarrow$} & \textbf{LPIPS$\downarrow$} & \textbf{User$\uparrow$} \\
      \hline
      StyleID & 28.8 & 18.1 & \textbf{0.50} & 8.2 \\  
      CAP-VSTNET & 31.2 & 18.7 & 0.61 & 6.7 \\  
      StyleFormer & 31.7 & 18.4 & 0.63 & 6.4 \\  
      IEContraAST & 30.4 & 18.8 & 0.53 & 7.3 \\  
      StyTR2 & 30.7 & 18.9 & 0.54 &  7.3 \\  
      InstantStyle & 34.6 & 20.2 & 0.63 & 8.0 \\  
      IP-Adapter & 35.7 & 20.9 & 0.63 & 7.9 \\  
      \hline
      \textbf{Ours} & \textbf{14.1} & \textbf{8.3} & 0.51 & \textbf{8.6} \\  
      \hline
    \end{tabular}
    \label{table:1}
  \end{center}
\end{table}

\subsubsection{Evaluation Metrics}
Like in StyleID~\cite{chung2024style}, we employ ArtFID~\cite{wright2022artfid}, FID~\cite{heusel2017gans}, and LPIPS~\cite{zhang2018unreasonable} as our metrics to ensure that both style similarity and structure fidelity are effectively measured. 
Since evaluating stylization is highly subjective, we also explore user preferences. We invite 50 testers, each of whom receives a file containing all stylization results in a randomized order. Testers rate each work, and we calculate the average score based on their evaluations.




\begin{figure*}[t]
    \centering
    \includegraphics[width=17.5cm]{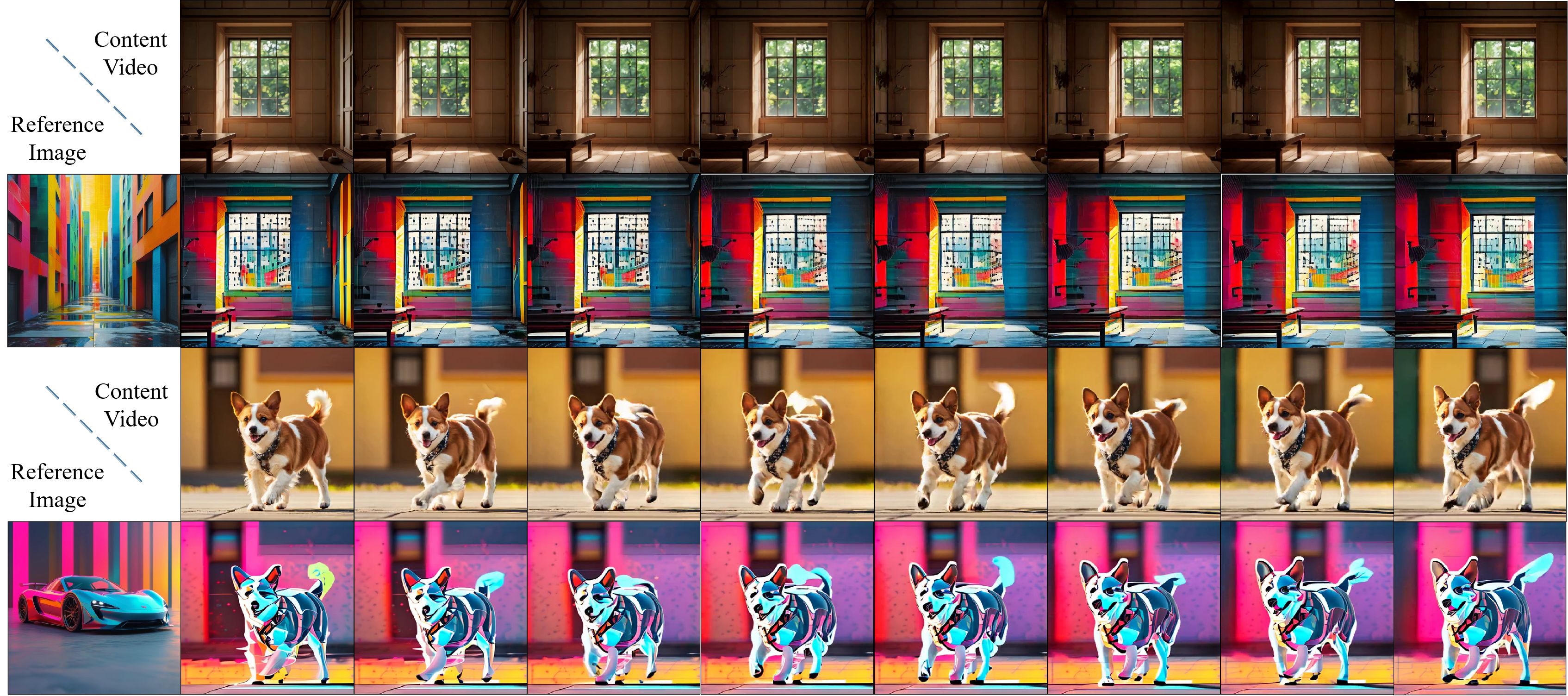}
    \caption{Results on videos. StyleBrush effectively achieves stylization even with video input, whether for static scene (with camera movement) or dynamic object (the dog). The results combines the structure of the contents and the style of reference images, and keep the inter-frame stability and continuity well.}
    \label{fig:video}
\end{figure*}

\subsection{Comparisons}

In this section, we compare our method with previous methods both qualitatively and quantitatively. 

\subsubsection{Qualitative Comparisons}

The proposed method has been compared with several highly influential works from recent years, including expert models like CAP-VSTNET \cite{wen2023cap}, IEContraAST \cite{ma2023rast}, StyleFormer \cite{wu2021styleformer}, and StyTR2 \cite{deng2022stytr2}. Additionally, we compared our approach with the latest training-free large model methods such as StyleID \cite{chung2024style}, as well as recent style transfer techniques like InstantStyle \cite{wang2024instantstyle} and IP-Adapter \cite{ye2023ip}. These works, which have demonstrated impressive stylization results, clearly represent the advancements in the three categories of stylization methods mentioned above. To ensure fairness, a qualitative comparison of the results was conducted under the same testing samples.

We show the visual comparison results in Figure~\ref{fig:4}. 
For IP-Adapter and InstantStyle (the last two columns), the tone and stroke of the style image are basically not maintained, and the results of the content image are seriously corrupted, which is reflected in almost all samples in the figure. For CAP-VSTNET, IEContraAst, and StyleFormer (column 4, 5, 6), the samples shown in the figure (e.g. in the first, second, fourth, fifth, and sixth rows) have problems such as messy strokes, overexposed colors, poor overall aesthetics, and many shadows. For StyleID (column 7), the degree of stylization is insufficient, as nearly all results fail to prominently display the details of the style. For StyleTR2 (column 8), images often contain some artifacts, and there is a problem with blurriness in the generated results (such as the shadows and artifacts in the second row, and the blurriness in the third and last rows). Overall, our results are the best in qualitative comparison.

\subsubsection{Quantitative Comparisons}

To more accurately demonstrate the reliability of our method, we conducted a quantitative analysis of the experimental results. We compared our proposed approach against state-of-the-art methods, including StyleID \cite{chung2024style}, 
CAP-VSTNET \cite{wen2023cap}, StyleFormer \cite{wu2021styleformer}, IEContraAST \cite{ma2023rast}, StyTR2 \cite{deng2022stytr2}, InstantStyle \cite{wang2024instantstyle}, and IP-Adapter \cite{ye2023ip}. The experimental results are shown in Table \ref{table:1}.
We conducted tests using a randomly selected set of $500$ image pairs. Our method achieved an ArtFID score of 14.1, an FID score of 8.3, and an LPIPS score of 0.51. These results demonstrate a significant improvement over previous work, with our method leading by 14.7 points in ArtFID and showing a substantial improvement of 9.8 points in FID compared to StyleID. Additionally, our LPIPS score is close to the state-of-the-art, with only a 0.1 difference. Note that our LPIPS score is comparable to that of StyleID, which means that our structure preservation ability is almost the same as StyleID, but our style degree is much higher than it, which is reflected in the ArtFID score and the visualization in Figure ~\ref{fig:4}. Furthermore, based on user preference scores, our method received an average score of 8.6, indicating that it was the most favored among all the tested approaches.
\subsection{Additional Analysis}

In this section, we conduct a study on the control of stylizaiton strength, which can be adjusted by users. Then, we test our results on videos and achieve stable and continuous stylization effects along frames.

\subsubsection{Study on Stylization Strength}
We visualized the results for style strength values of 0, 0.2, 0.4, 0.6, 0.8, and 1.0, as shown in Figure~\ref{fig:4}. It can be observed that when the strength is low, the image structure remains almost identical to the input. As the strength increases, the stylization effect becomes more pronounced, with the result at strength = 1.0 representing the fully stylized outcome. This property allows users to adjust the degree of stylization to suit their needs.

\subsubsection{Results on Video}
The StyleBrush technique excels in video stylization. As shown in the Figure \ref{fig:video}, our method effectively transforms the scene from the original wood color to a style similar to the reference image. Moreover, in the example of the running dog, the stylized result inherits the color and texture of the sports car well. Thanks to the motion module of the Animatediff architecture~\cite{guo2023animatediff}, we ensure overall video stylizaiton while maintaining the stability and continuity across frames, which shows great video stylization effect of our framework.

\section{Conclusion}
In conclusion, our study successfully developed \textbf{\papername}, a novel stylization method that efficiently harnesses styles from reference images and applies these styles to various visual content. Throughout the project, we employed a dual-branch architecture comprising ReferenceNet and Structure Guider. ReferenceNet was tasked with extracting stylistic elements from reference images, while Structure Guider focused on retaining the structural integrity of the input visuals, thus facilitating precise image-guided stylization. Our training approach was innovative, utilizing LLM and T2I models to dynamically generate style images. By cropping different areas of a single style image, we effectively constructed low-cost training pairs. Ultimately, the experimental results confirmed that our method attained state-of-the-art performance, as evidenced by both qualitative and quantitative evaluations. In future work, we aim to delve deeper into the expressive potential of art and enhance the integration of artistic concepts with AI technologies.

\bibliography{aaai24}

\end{document}